\documentclass[conference]{IEEEtran}
\usepackage{times}

\usepackage[numbers]{natbib}
\usepackage{multicol}
\usepackage[bookmarks=true]{hyperref}
\usepackage{todonotes}
\usepackage{xcolor}
\usepackage{graphicx}
\usepackage{amsmath}
\usepackage{amsfonts}
\usepackage{algorithm}
\usepackage{algpseudocode}
\usepackage{subcaption}
\usepackage{booktabs}

\begin{document}

\title{Curriculum Reinforcement Learning\\for Complex Reward Functions}




%
\author{
\authorblockN{
Kilian Freitag\textsuperscript{1}, 
Kristian Ceder\textsuperscript{1}, 
Rita Laezza\textsuperscript{1}, 
Knut Åkesson\textsuperscript{1}, 
Morteza Haghir Chehreghani\textsuperscript{1,2}
}
\authorblockA{\textsuperscript{1}Chalmers University of Technology\\
Gothenburg, Sweden
}
\authorblockA{\textsuperscript{2}University of Gothenburg\\
Gothenburg, Sweden\\
Email: \{tamino, cederk, laezza, knut.akesson, morteza.chehreghani\}@chalmers.se
}
}

\maketitle

\begin{abstract}
Reinforcement learning (RL) has emerged as a powerful tool for tackling control problems, but its practical application is often hindered by the complexity arising from intricate reward functions with multiple terms. The reward hypothesis posits that any objective can be encapsulated in a scalar reward function, yet balancing individual, potentially adversarial, reward terms without exploitation remains challenging. To overcome the limitations of traditional RL methods, which often require precise balancing of competing reward terms, we propose a two-stage reward curriculum that first maximizes a simple reward function and then transitions to the full, complex reward. We provide a method based on how well an actor fits a critic to automatically determine the transition point between the two stages. Additionally, we introduce a flexible replay buffer that enables efficient phase transfer by reusing samples from one stage in the next. We evaluate our method on the DeepMind control suite, modified to include an additional constraint term in the reward definitions. We further evaluate our method in a mobile robot scenario with even more competing reward terms. In both settings, our two-stage reward curriculum achieves a substantial improvement in performance compared to a baseline trained without curriculum. Instead of exploiting the constraint term in the reward, it is able to learn policies that balance task completion and constraint satisfaction. Our results demonstrate the potential of two-stage reward curricula for efficient and stable RL in environments with complex rewards, paving the way for more robust and adaptable robotic systems in real-world applications.
\end{abstract}

\IEEEpeerreviewmaketitle

\section{Introduction}
Reinforcement Learning (RL) has emerged as a powerful paradigm in the field of robotic control, offering the promise of adaptable and efficient solutions to complex problems. RL has demonstrated its potential to learn optimal policies in a wide range of applications such as manipulation \citep{gu2016deep, kalashnikov2018scalable, leyendecker2022deep, han2023survey} or mobile robotics \citep{lillicrap2015continuous, zhu2017target, rodriguez2018deep, zhu2021deep, sang2022motion}. However, as we transition from carefully curated benchmarks to realistic scenarios, a significant gap emerges, highlighting the challenges of applying RL in practical settings.

One of the primary challenges in realistic applications lies in the complexity of the environment and the multiplicity of objectives. While classical RL problems often focus on a single, well-defined goal, real-world scenarios typically involve multiple, sometimes conflicting objectives. For instance, a mobile robot might need to navigate to a goal location while simultaneously avoiding obstacles, maintaining a specific velocity, and ensuring a smooth trajectory \citep{zhang2023collision, ceder2024traj}. This multi-objective nature of real-world problems poses a significant challenge to traditional RL approaches. 

The reward hypothesis suggests that any learning objective can be expressed by a single scalar reward under certain assumptions \citep{sutton2018reinforcement, bowling2023settling}. However, formulating an effective reward function for complex tasks is non-trivial and can often result in undesired behaviors \citep{booth2023perils, knox2023reward, knoxspecify}. Moreover, optimizing such complex rewards can be challenging due to the presence of local optima, where policies might satisfy only a subset of objectives (e.g., minimizing energy consumption by remaining stationary) without considering others. We refer to such potentially conflicting objectives as complex reward functions, where the complexity lies in the presence of strong local optima for undesired behaviors.

A line of work that has emerged to tackle challenging RL problems is curriculum learning \citep{bengio2009curriculum,narvekar2020curriculum}. Inspired by how animals can be trained to learn new skills \citep{skinner1958reinforcement}, learning proceeds from simple problems to gradually more complex ones. In the realm of RL, such curricula are often hand-designed and have been successfully employed in several applications in robotic control \citep{sanger1994neural, hwangbo2019learning, leyendecker2022deep}.
More recently several methods for automatic curriculum design have emerged that are able to learn curriculum policies \citep{narvekar2019learning} or that break down problems into smaller ones \citep{andrychowicz2017hindsight, fang2019curriculum} for more effective learning. Further, automatic curricula are used in sparse or no reward settings as intrinsically motivated exploration that encourages reaching diverse states \citep{bellemare2016unifying, pathak2017curiosity, burda2018exploration, shyam2019model}. Nevertheless, such methods have been explored less for complex reward functions.

\begin{figure*}[t]
  \centering
  \includegraphics[width=\textwidth]{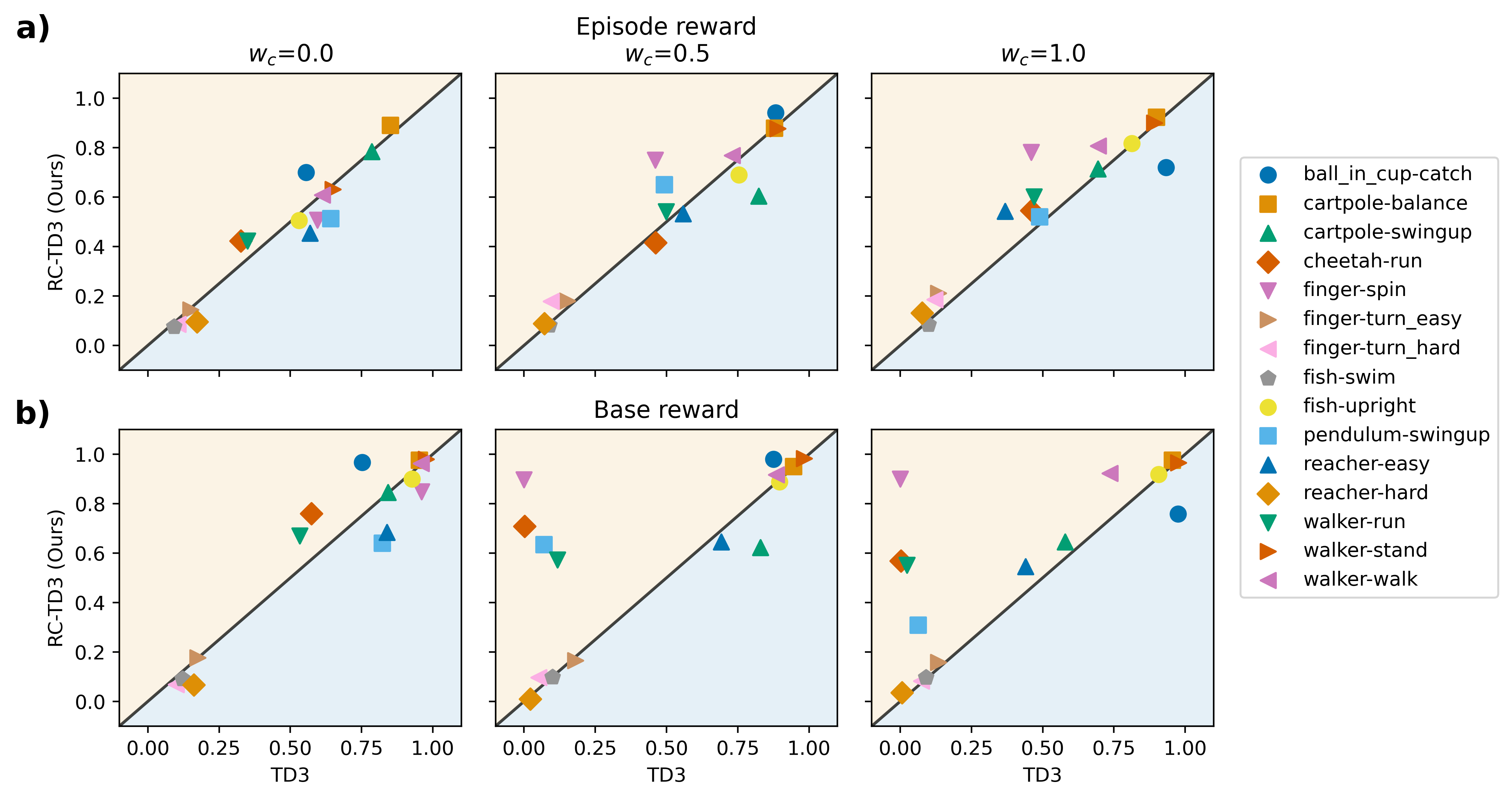}
  \caption{Comparison of the standard TD3 algorithm with reward curriculum version (RC-TD3). Fig. a) shows the normalized mean episode reward visualized using constraint weight $w_c=1.0$ for intuitive comparability between policies trained on different constraint weights $w_c$. As it can be seen, the curriculum becomes more effective with higher $w_c$. It seems to be most helpful for environments where the constraints are not automatically optimized by completing the task, i.e. the ones in the center of the left plot with $w_c=0.0$. In environments where learning is unsuccessful in the first place or that are relatively simple even with constraints the effects of the curriculum are less pronounced. Fig. b) shows the mean base reward $r_b$ without constraints. Importantly, employing a reward curriculum manages to keep or improve the base reward in almost all cases, especially for high $w_c$. This demonstrates its effectiveness in finding a better trade-off between task performance and constraint satisfaction, given that the baseline often gets stuck in the local minima of only optimizing constraints as in the case of finger spin.}
  \label{fig:td3_improve}
\end{figure*}

To address the challenge of learning with complex reward functions, we propose leveraging curriculum learning. We introduce a novel two-stage reward curriculum combined with a flexible replay buffer to effectively balance task success and constraint satisfaction. In the first phase, a subset of rewards is used for training to simplify the discovery of successful trajectories. When the policy has converged sufficiently, the second phase is initiated, optimizing the full reward. To automatically determine when to switch to the second phase, we track how well the actor optimizes the $Q$-function as a proxy for policy convergence. Additionally, our method allows for sample-efficient reuse of collected trajectories by incorporating two rewards in the replay buffer such that samples from the first phase can be reused for training with an updated reward in the second phase.

To analyze the efficacy of our method, we first evaluate it on several modified environments from the DeepMind control suite, where the agent in addition to the original reward gets penalized for taking large actions. Furthermore, we evaluate it on a mobile robot that is tasked to reach a goal location while avoiding collisions. Simultaneously, it must fulfill several constraints, including maintaining a reference velocity, staying close to a planned path when possible, and generating smooth trajectories. We formulate the reward in an intuitive way, to enable simple testing of different constraint weights $w_c$. The complexity again does not lie in finding any valid solution, but in finding a solution that maximally satisfies constraints while still reaching the goal, exacerbated by the plurality of the objective.

In our experiments, we compare our method to a baseline that always trains on the full reward. We show that our two-stage reward curriculum becomes more effective the higher the constraints are weighted in the overall reward. Especially for environments where the constraints hinder solving the task, our method is able to successfully learn the task while satisfying constraints where the baseline fails and gets stuck in solely optimizing constraints (see Fig. \ref{fig:td3_improve}). In an ablation study where we either reset the network weights or the replay buffer when changing curriculum phases, we show that the main factor for the success of the method lies in the "pretrained" network weights from the first phase. Resetting the replay buffer does not greatly change the outcome. Furthermore, we examine how our automatic curriculum switch performs in comparison to a static switch after a predetermined number of steps. Our results demonstrate that the automatic switch effectively identifies suitable switching times, leading to faster convergence than the static approach.
Our contributions can be summarized as follows:

\begin{enumerate}
    \item We introduce a novel two-stage reward curriculum to effectively learn complex rewards by reusing past experiences with updated rewards and an automatic mechanism to switch phases. The curriculum is integrated in two RL methods, one based on SAC and the other based on TD3.
    \item We extensively evaluate our method on several modified DeepMind control suite environments and discuss in which environments a reward curriculum is most effective.
    \item In ablation studies we demonstrate that the key ingredients for our method are the network weights obtained in the first phase and the automatic switching mechanism.
    \item To understand the impact of our method on environments with several conflicting reward terms, we extensively evaluate it on a mobile robot navigation problem.
\end{enumerate} 

We will make the code used for our experiments openly available.

\section{Problem Formulation}\label{sec:env}
We formulate the problem as a Markov Decision Process (MDP) defined by the tuple $\langle\mathcal{S}, \mathcal{A}, \mathcal{P}, r, p_0, \gamma\rangle$. $\mathcal{A}$ represents the action space, $\mathcal{S}$ the state space, $\mathcal{P}:\mathcal{S}\times\mathcal{A}\times\mathcal{S}\rightarrow[0,1]$ is the state transition function, $p_0:\mathcal{S}\rightarrow[0,1]$ the initial state distribution and $\gamma\in[0,1)$ a discount factor. The reward is denoted by $r$; we omit its explicit dependence on state and action $r(s,a)$ for a less clustered notation.

The objective of RL is to maximize the expected return, $\mathbb{E}[G_n]$, where $G_n = \sum_{k=n}^{N} \gamma^{k-n} r_k$ is the cumulative discounted reward, with $n$ being the current step and $N$ the maximum steps per episode. A key concept for achieving this goal is the action-value function, commonly referred to as the $Q$-function, which represents the expected return when taking action $a$ in state $s$. The Q-function is typically parameterized by $\phi$, and denoted $Q_\phi(s,a)$.

In our case, we consider any control problem with several reward terms, where each can be categorized either as a base reward $r_b$ if it helps learning the goal (for instance reward shaping terms), and constraint rewards $r_c$, which specify the desired behavior. The reward is then given as the weighted sum
\begin{align}
    r = r_b + w_c r_c
\end{align}\label{eq:reward_cr}

Given this problem formulation, the goal is to develop methods that allow learning a policy $\pi(a|s)$ that learns to complete tasks while maximally satisfying the constraints $r_c$ and being robust to different constraint weights $w_c$. Intuitively, the challenge lies in finding policies that learn the task without exploiting rewards that relate to the constraints.

\section{Reward Curriculum}\label{sec:reward_cr}

\begin{algorithm}
\caption{Off-policy Reward Curriculum}\label{alg:crsac}
\begin{algorithmic}
\State Initialize network parameter $\phi_1, \phi_2, \bar{\phi}_1, \bar{\phi}_2, \theta, \bar{\theta}$
\State $t \gets 0$
\State $\mathcal{CR} = 0$ \Comment{Initial curriculum phase}
\For{each iteration}
    \For{each environment steps} 
        \If{using SAC}
            \State $a_t \sim \pi_\theta(a_t|s_t)$
        \Else
            \State $a_t = \pi_\theta(s_t) + \epsilon \quad \epsilon \sim \mathcal{N}(0,\sigma)$
        \EndIf
        \State $s_{t+1} \sim p(s_{t+1}|s_t,a_t)$
        \State $\mathcal{D} \gets \mathcal{D} \cup \{(s_t, a_t, r_b, r, s_{t+1})\}$
        \State $t \gets t + 1$
    \EndFor
    \If{$\forall i \in \{t-m+1,\ldots,t\}: J_{\pi,Q;i} < J_{\pi,Q;CR}$}
        \State $\mathcal{CR} = 1$
    \EndIf
    \For{each gradient steps} 
        \State $B = \{(s, a, r_{b}, r, s')\} \sim \mathcal{D}$ 
        \If{$\mathcal{CR} == 0$}  \Comment{Choose reward} 
            \State $r_{\text{cr}} \gets r_{b}$
        \Else
            \State $r_{\text{cr}} \gets r$
        \EndIf

        \State $B \gets \{(s, a, r_{\text{cr}}, s')$ 

        \For{$i \in \{1,2\}$} 
            \State $\phi_i \gets \phi_i - \lambda_Q \nabla_{\phi_i} \frac{1}{|B|} \sum_{(s,a,r_{\text{cr}},s')\in B} J_Q(\phi_i)$
        \EndFor
        \State $\theta \gets \theta - \lambda_\pi \nabla_\theta \frac{1}{|B|} \sum_{s\in B} J_\pi(\theta)$ \Comment{Delayed in TD3}

        \If{using SAC}
            \State $\alpha \gets \lambda_\alpha \nabla_\alpha J(\alpha)$
        \EndIf
        \State $\bar{\phi}_i \gets \tau_{\text{targ}} \bar{\phi}_i + (1-\tau_{\text{targ}})\phi_i \ \text{for} \ i \in \{1,2\}$

        \If{using TD3}
            \State $\bar{\theta} \gets \tau_{\text{targ}} \bar{\theta} + (1-\tau_{\text{targ}})\theta$
        \EndIf
    \EndFor
\EndFor
\end{algorithmic}
\end{algorithm}

We propose a novel two-stage reward curriculum, to effectively learn complex reward functions in a sample-efficient manner. In principle the reward curriculum can be combined with any off-policy RL algorithm, though in this work we focus on two versions of our method: one based on Soft-Actor Critic (SAC) \citep{haarnoja2018soft} and the other based on Twin-Delayed DDPG (TD3) \citep{fujimoto2018addressing}, which we denote as RC-SAC and RC-TD3 respectively. In the first phase of the curriculum, we consider only a subset of reward terms for training, denoted as $r_{b}$. In the second phase, we then directly optimize the full reward $r=r_b+w_c r_c$. Notably, we store the rewards from both phases in the replay buffer, resulting in the tuple $\{( s_t, a_t, r_{b}, r, s_{t+1} )\}$. This allows us to reuse past experiences when transitioning to the second phase, where we switch the reward during training to ensure stable and sample-efficient learning.

Formally, we define the curriculum reward as
\begin{align}\label{eq:cr_reward}
    r_{\text{cr}} = \begin{cases}
            r_{b} & \text{if} \quad \mathcal{CR} = 0 \\
            r & \text{otherwise}
       \end{cases}
\end{align}
where $\mathcal{CR}$ is the current curriculum phase. Algorithm \ref{alg:crsac} describes the two-stage curriculum in detail.

\subsection{RC-SAC}
In the following, we show how our method integrates with SAC. Instead of solely optimizing the expected return, SAC makes use of the maximum entropy objective \citep{ziebart2010modeling} such that a policy additionally tries to maximize its entropy $\mathcal{H}$ at each state which is defined as
\begin{align*}
    \pi_\theta^* = \text{arg} \max_{\pi_\theta} \sum_t \mathbb{E}_{(s_t,a_t)\sim \rho_{\pi_\theta}} \left[r_{\text{cr}}(s_t,a_t) + \alpha \mathcal{H}(\pi_\theta(\cdot|s_t) \right]
\end{align*}
where $\alpha$ is a parameter to control the policy temperature, i.e. the relative importance of the entropy term. SAC makes use of two Q-functions parameterized by $\phi_1$ and $\phi_2$ which are updated as
\begin{align}
    J_Q(\phi_i) = (Q_{\phi_i}(s,a)-y(r_{cr},s'))^2
\end{align}\label{eq:q_loss}
where $i$ is the Q-function index and the targets are computed as
\begin{align*}
    y(r_{cr},s') = r_{\text{cr}} + \gamma \left[\min_{i=1,2} Q_{\bar{\phi}_i}(s', \tilde{a}') - \alpha \log{\pi_\theta(\tilde{a}'|s')}\right]
\end{align*}
Importantly, $\tilde{a}'\sim\pi_\theta(\cdot|s')$ is newly sampled during training and $Q_{\text{targ},i}(s, a)$ is the $i$th target Q-function. Note that instead of optimizing for a general reward $r$, we use $r_{\text{cr}}$ in this step which changes depending on the phase. The policy update is given by
\begin{align*}
    J_\pi(\theta) = \min_{i=1,2}Q_{\phi_i}(s,\tilde{a}_\theta(s)) - \alpha \log{\pi_\theta(\tilde{a}_\theta(s)|s)}
\end{align*}\label{eq:sac_actor_loss}
where $\tilde{a}_\theta(s)$ is sampled from $\pi_\theta(\cdot|s)$ via the reparameterization trick \citep{ruiz2016generalized}.
We make use of the entropy-constraint variant of SAC as described in \citep{haarnoja2018soft}, where $\alpha$ is updated as
\begin{align*}
    J(\alpha)=\mathbb{E}_{a\sim\pi_\theta} \left[-\alpha \log{\pi_\theta(a|s)} - \alpha \mathcal{H}(\pi_\theta(\cdot|s)) \right]
\end{align*}
An important parameter in SAC is the initial value for $\alpha$, which we denote as $\alpha_{\text{init}}$. It determines the importance of the entropy term and, thus how much the agent explores different states. When $\alpha$ converges to low values, the agent focuses mainly on the objective instead and becomes more deterministic. For our experiments, we set $\alpha_{\text{init}}=1.0$ and clip it to a minimum of $\alpha_{\text{min}}=0.0001$ to increase stability.

\begin{figure}[t]
  \centering
  \includegraphics[width=\columnwidth]{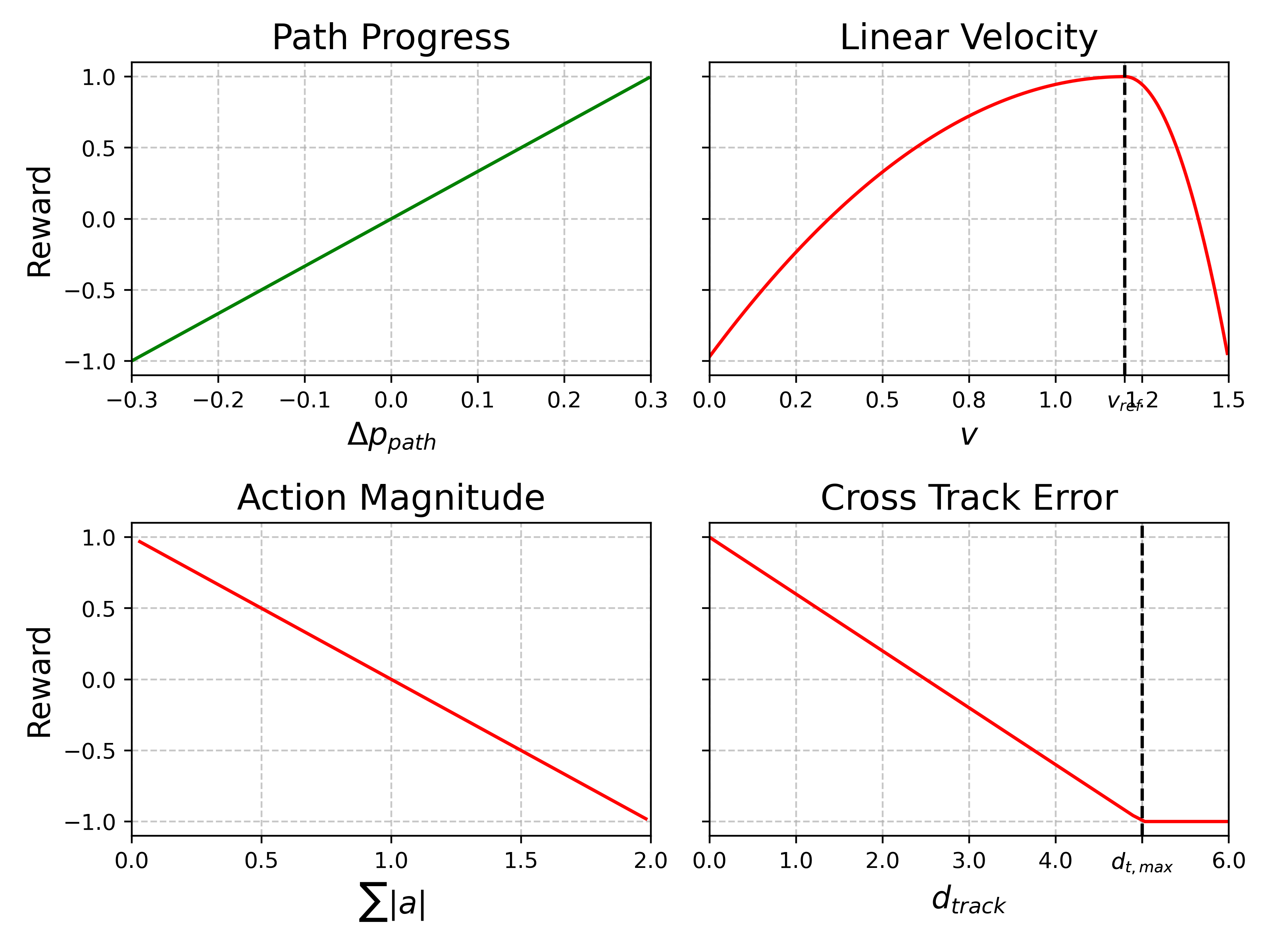}
  \caption{Functions for dense reward terms with $\kappa=0.942$, $v_{\text{ref}}=1.2$ and $d_{\text{track,max}}=5$. The range for each term is normalized to $[-1, 1]$. Green shows the reward shaping term that enables finding the goal faster. The soft constraints are colored in red.}
  \label{fig:dense_reward_functions}
\end{figure}

\subsection{RC-TD3}
Furthermore, we show how the reward curriculum integrates with Twin-Delayed DDPG (TD3) \citep{fujimoto2018addressing}. It extends the Deep Deterministic Policy Gradient (DDPG) \citep{lillicrap2015continuous} algorithm to enhance its stability by making use of two Q-functions, delay the policy updates, and add noise to target actions to avoid exploitation of Q-function errors.

It uses a deterministic policy, though for sampling an action Gaussian noise is added for improved exploration
\begin{align*}
    a_t = \pi_\theta(s_t) + \epsilon \quad \epsilon \sim \mathcal{N}(0,\sigma)
\end{align*}
A similar procedure but with clipped noise is used when calculating the next action to update the estimated Q function to avoid exploitation of overestimations as
\begin{align*}
    \tilde{a} = \pi_{\bar{\theta}}(s') + \epsilon \quad \epsilon \sim \text{clip}(\mathcal{N}(0,\tilde{\sigma}),-c,c)
\end{align*}
where $\pi_{\bar{\theta}}$ is a target policy.
The Q targets are then computed as
\begin{align*}
    y(r_{cr},s') = r_{cr} + \gamma \min_{i=1,2} Q_{\bar{\phi}_i}(s', \tilde{a}')
\end{align*}
and the Q function is updated as in Equation \ref{eq:q_loss}.
The policy is updated using the first Q network as
\begin{align}
    J_\pi(\theta) = Q_{\phi_1}(s,\pi_\theta(s))
\end{align}\label{eq:td3_actor_loss}
Importantly, the policy is not updated in every iteration. Often it is delayed to update only in every second iteration, which we also employ in this work. In all experiments we set $\tilde{\sigma}=0.2$, $c=0.5$ and $\sigma=0.1$. For more details about the parameters and specific design choices, we refer to the original paper.

\begin{figure}[t]
  \centering
  \includegraphics[width=\columnwidth]{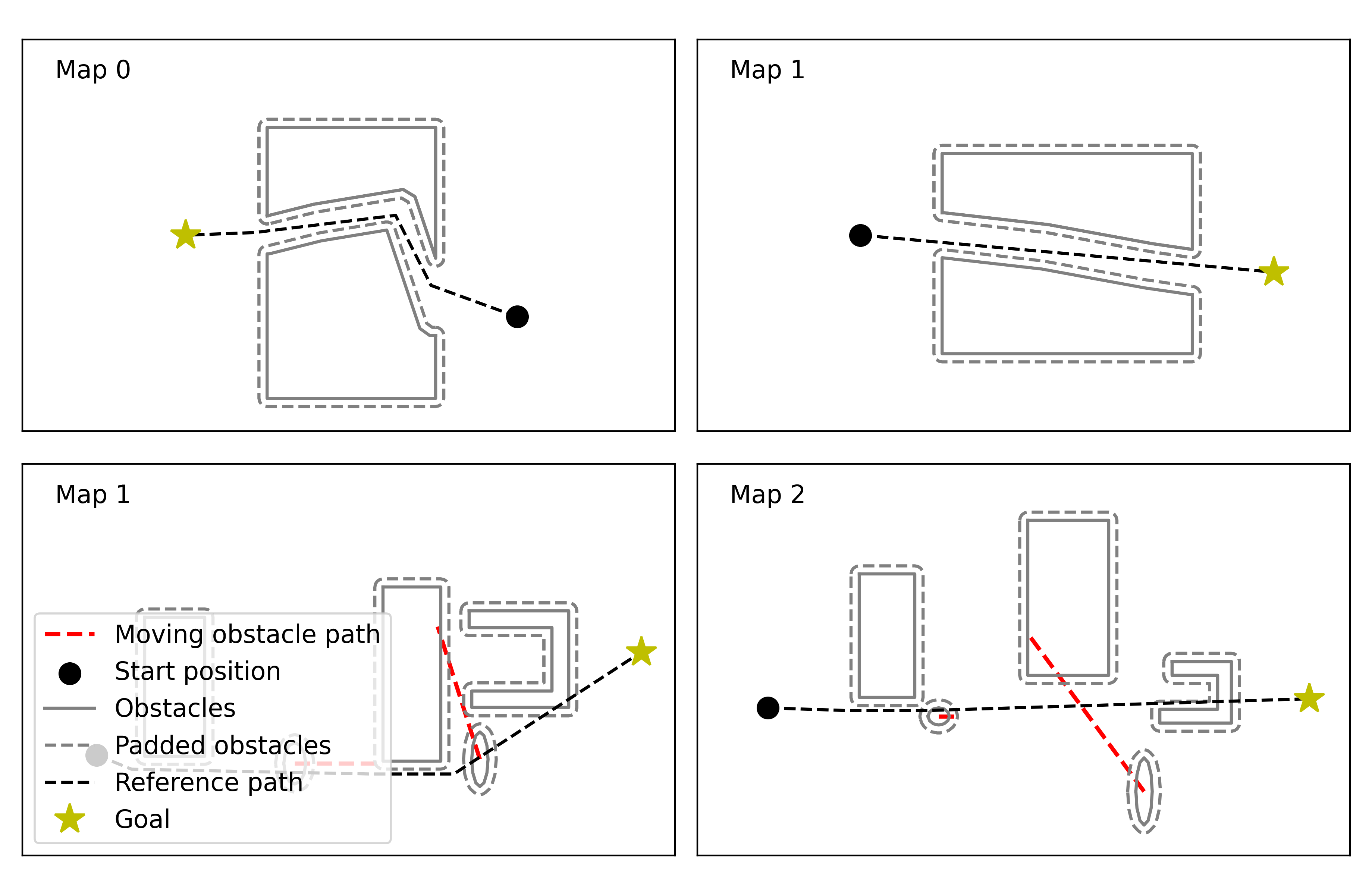}
  \caption{Exemplary environment maps used for training. Obstacle positions, paths, initial states, and goal positions are randomized. While maps 0 and 2 contain dynamic obstacles, maps 1 and 3 only contain static ones.}
  \label{fig:maps}
\end{figure}

\subsection{Automatic phase switch}\label{sec:autoswitch}
One key aspect of the reward curriculum is when to switch from the initial phase to training with the full reward. We hypothesize that the actor has to sufficiently learn the task using the base reward before constraints should be added. As a proxy for learning, we consider how well the actor optimizes the critic. In the case of TD3 that would simply be the normal actor loss \ref{eq:td3_actor_loss}, for SAC it would be the actor loss \ref{eq:sac_actor_loss} but without the additional entropy term. If the actor fits the critic well for $m$ timesteps, i.e. $\forall i \in \{t-k+1,\ldots,t\}: J_{\pi,Q;i} < \Gamma_{\text{CR}}$ where $\Gamma_{\text{CR}}$ is the threshold that defines a "good" fit, we switch to the second curriculum phase.

We also considered other metrics to find when to best initiate the second phase such as policy entropy, but found that it is not straightforward to relate entropy to policy convergence, and in addition this would limit the algorithm to stochastic policies. Further, we hypothesize that the Q fit would be a more stable and accurate measure compared to simply tracking the obtained reward.

\section{Environments}
We first evaluate the methods RC-SAC and RC-TD3 on several different environments from the DeepMind control suite and then on a mobile robot environment with several reward terms. In the following, we describe the details of the environments.

\subsection{DeepMind Control Suite}
The DeepMind Control Suite \citep{tassa2018deepmind} is a well-known collection of control environments in RL. We use the state-space observations and the environments have an action space $\mathcal{A} \in [-1,1]^2$. The original rewards are always in $r\in [0,1]$, which we take as base reward $r_b$. Furthermore, we introduce the following reward term to minimize action magnitudes, to find efficient solutions to the control problems
\begin{align}
    r_c=-\frac{1}{d}\|a\|_1
\end{align}
which is used as the constraint term in the curriculum reward \ref{eq:reward_cr}.

\subsection{Mobile Robot}
To evaluate our method in a more realistic and complex setting (in terms of rewards), we consider a mobile robot navigation problem where the robot is tasked to reach a goal position while avoiding obstacles and satisfying various constraints. The environment maps are randomized and contain both permanent (e.g., walls and corridors) and temporary (e.g., dynamic or static) obstacles. A subset of exemplary environment maps can be found in Figure \ref{fig:maps}. Furthermore, a reference path is computed using A* \cite{Hart1968}, considering only permanent obstacles. This reference path should simulate a setting where an optimal path is pre-computed but the robot might not be able to naively follow it due to temporary obstacles. The constraints include driving at a reference velocity, staying close to the reference path, and creating smooth trajectories. 

The state space $\mathcal{S} \in \mathbb{R}^{178}$ consists of the latest two lidar observations, the robot's position, the current speed, the reference path, and the goal position.
The action space $\mathcal{A} \in [-1,1]^2$ represents the translational and angular acceleration. 

For the reward design, we will focus on easily interpretable formulations. There are three possible outcomes of an episode: (1) reach the goal, (2) timeout, i.e. reach maximum steps, or (3) collide with an obstacle. Empirical tests have shown that penalizing collisions mainly hinders exploration and does not lead to better final policies. Therefore, the only outcome-based reward included is:
\begin{align*}
\begin{split}
    r_g &= \begin{cases}
            100 & \text{if reached goal}\\
            0 & \text{otherwise}
           \end{cases}
\end{split}
\end{align*}

\begin{figure*}[t]
  \centering
  \includegraphics[width=\textwidth]{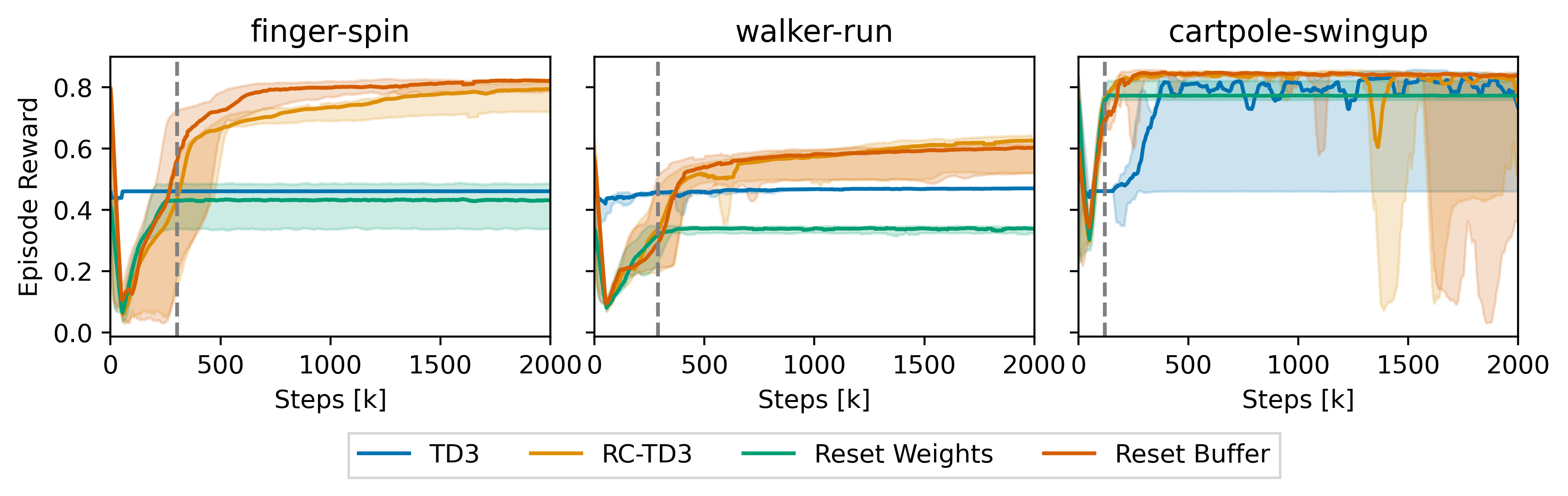}
  \caption{Investigation of the median episode reward of RC-TD3 to resetting the network weights and resetting the networks when changing curriculum phases. The values are smoothed by taking the running average with window size 50 [k]. As it can be seen, resetting the network deteriorates performances while resetting the replay buffer has relatively little influence on the final outcomes. We conclude that the main benefit of a reward curriculum in these environments comes from improved exploration given by a ``pretrained'' network. The gray dashed line indicated the mean time when the curriculum phase was switched in RC-TD3. }
  \label{fig:td3_resets}
\end{figure*}

The other objectives in the mobile robot problem can be expressed as soft constraints evaluated at each step, which provide dense reward signals. To enable intuitive weighting, we normalize each term such that all soft constraint rewards are between $[-1,1]$. We chose this range over $[0,1]$ to discourage the robot from learning to immediately crash.

We consider minimizing accelerations to achieve smooth trajectories. As this corresponds to our action space, we penalize high action values, similar to the other experiments, as
\begin{align*}
    r_{a} = 1 - \sum_{i=1}^2 |a_{i}|, \ \  \ r_{a} \in [-1, 1]
\end{align*}

To encourage driving at a desired reference velocity $v_{\text{ref}}$, we model a velocity reward as a piece-wise quadratic function centered on $v_{\text{ref}}$, and scaled to our desired range. This choice is motivated by the intuition that slowing down is less severe than going too fast. 
\begin{align*}
    r_{v} = 1 - \frac{l_2^\kappa(v_t-v_{ref}) \cdot 2}{max(l_2^\kappa(-v_{ref}), l_2^\kappa(v_{max}-v_{ref}))},   \ \  r_{v}  \in [-1, 1]
\end{align*}
with the piece-wise quadratic function $l_2^\kappa$ being defined as:
\begin{equation*}
l_2^\kappa(x) = \begin{cases}
    \kappa x^2 & \text{if } x>0 \\
    (1-\kappa) x^2 & \text{otherwise}
\end{cases}
\end{equation*}
where $\kappa$ controls the slope of the negative and positive regions and $v_{max}=1.5$ is the maximum velocity.

Further, we linearly penalize deviating from the reference trajectory with the following reward term
\begin{equation*}
r_{x} = \text{clip}\left(\frac{|d_{\mathit{track}}|}{d_{\mathit{track},\max}}, -1, 1\right), \ \ r_{x} \in [-1, 1]
\end{equation*}
where $d_{\mathit{track}}$ is the distance between the agent and the reference trajectory and $d_{\mathit{track},\max}$ is a tuning parameter that sets the maximum distance. This is done to make the robot's behavior predictable, such that it only deviates from a planned path if necessary. 

To enable effective learning we additionally make use of potential-based reward shaping \cite{ng1999policy}, by encouraging progress along the reference path, through
\begin{align*}
    r_p = \frac{p_{\text{path}}(s') - p_{\text{path}}(s)}{v_{\text{max}} \cdot dt}, \ \ r_p \in [-1, 1]
\end{align*}
where $p_{\text{path}}(s)$ is the position on a path in state $s$ and the denominator is the maximum distance traversed in one step. As it is potential-based, it does not alter the ordering over policies \cite{ng1999policy}. Figure \ref{fig:dense_reward_functions} gives an overview of the dense reward terms used.

Assuming that all constraints are equally important we assign equal constraint weights $w_c$. The reward is then given by
\begin{align}\label{eq:true_reward}
    r = r_{g} + w_p r_p + w_c (r_{v} + r_{a} + r_x)
\end{align}
where we set the reward-shaping weight $w_p=0.25$ throughout all experiments.

\subsection{Experimental setup}

As neural network architecture, we use two fully connected layers, each with 256 hidden units and ReLU activation. We use a replay ratio of 1, where we first sample $1000$ environment steps and then train for the same number of gradient steps. Our replay buffer has a capacity of $1_000\,000$ samples and we set $\lambda_Q=\lambda_\pi=\lambda_\alpha=3.0\times 10^{-4}$, $\tau_{\text{targ}}=0.995$, and the batch size to 128.
In the DeepMind control suite experiments we set $\Gamma_{\text{CR}}=-50$ with $m=20$ (in [k]) for both RC-TD3 and RC-SAC, while we use $\gamma=0.999$ for RC-TD3 and $\gamma=0.99$ for RC-SAC with a maximum of $1000$ steps per episode.
For the mobile robot we use $\gamma=0.99$, $m=20$, $\Gamma_{\text{CR}}=-6$ for RC-TD3 and $\Gamma_{\text{CR}}=-20$ for RC-SAC with a maximum of $300$ steps per episode.
We utilized Nvidia T4 and A40 GPUs as computational resources. 
All results are computed using 4 random seeds. Further, we report the rewards (unless stated otherwise) with the fixed values $w_c=1.0$ for DM Control and $w_c=0.1$ for the mobile robot case, even though $w_c$ might have different values during training. We do this in order to make runs with different $w_c$ comparable. We report the normalize episode reward unless stated otherwise, such that 0 is the worst and 1 is the best.

\section{Results \& Discussion}

\begin{figure*}[t]
  \centering
  \includegraphics[width=\textwidth]{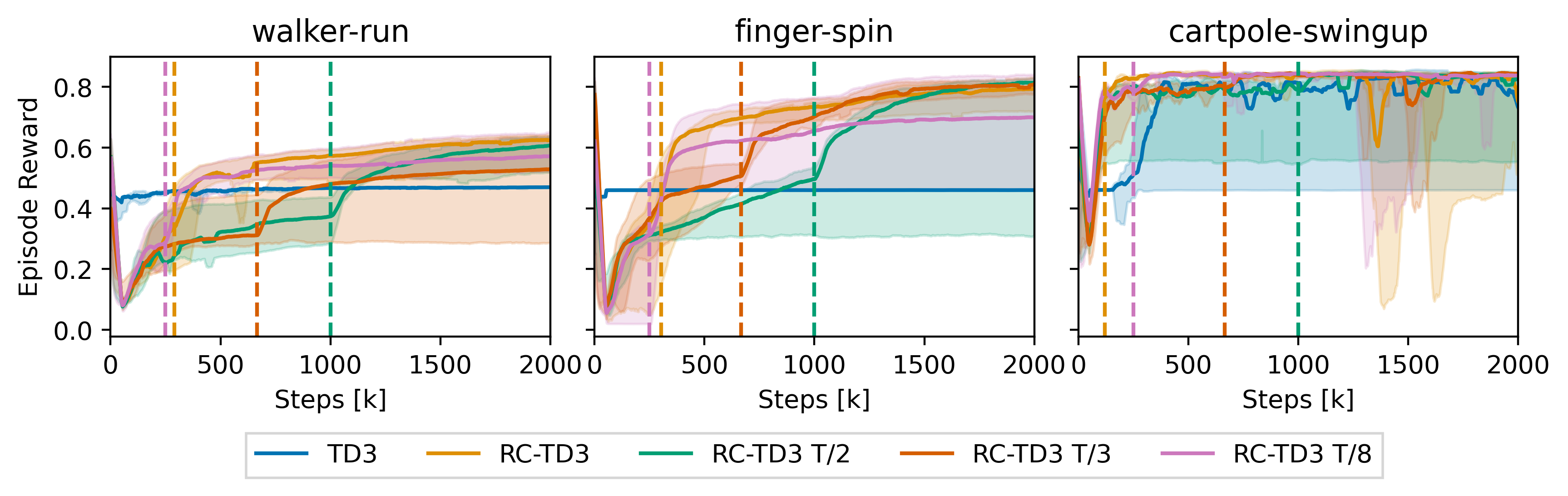}
  \caption{Comparison of the median episode reward of RC-TD3 with automatic curriculum switch as described in Section \ref{sec:autoswitch} to switching at times T/8, T/3, and T/2 where T are the total number of timesteps. The values are smoothed by taking the running average with window size 50 [k] and the dashed lines indicate the time each curriculum switched phases. The results indicate that for some environments such as cartpole swingup, the exact time does not seem to matter while the automatic switch manages to pick a more beneficial time in the case of walker run. Importantly, all reward curriculum versions work better than not employing a curriculum and our method, RC-TD3, yields one of the best results in all cases. }
  \label{fig:td3_switch}
\end{figure*}

\subsection{DM Control}
We test our method for $w_c\in \{0.0, 0.5, 1.0\}$ on the DeepMind Control Suite environments pendulum swingup, fish upright, fish swim, reacher hard, reacher easy, finger turn hard, finger turn easy, finger spin, ball in cup catch, cartpole balance, cheetah run, cartpole swingup, walker stand, walker walk and walker run. We train for $2,000,000$ steps. In Fig. \ref{fig:td3_improve} a) we show the mean episode reward (calculated with $w_c=1.0$ for comparability) obtained with RC-TD3 on the Y-axis versus the episode reward with TD3 on the X-Axis. A table with mean values and standard deviations can be found in the Appendix \ref{tab:td3_dm}. An X above the diagonal indicates that the curriculum improves performance, and below implies the opposite. The results are averaged over the last $50,000$ steps and 4 random seeds. As expected, both methods perform similarly when $w_c=0.0$. However, when $w_c$ increases, the curriculum becomes more effective. This is especially the case for environments that do not automatically optimize the constraint by completing the task. For instance, in cartpole balance it is beneficial for the task to predict small actions, thus a curriculum does not have a sizable effect. However, for tasks like finger spin, adding the constraints seems to hinder learning the task, leading to suboptimal policies with TD3 as $w_c$ increases. In those cases, where the task can be learned but constraints substantially impact learnability, our curriculum method proves to be most effective (i.e. environments with episode reward around the range from -100 to 500). Furthermore, there is one group in the bottom left, where the policy does not successfully learn the problem without constraints. In this case, the curriculum does not have any effect and learning remains unsuccessful. In Fig. \ref{fig:td3_improve} b) we show the base reward $r_b$, thus the reward that only encodes the task without any constraints for both TD3 and RC-TD3. As can be seen, the reward curriculum becomes more beneficial as the constraint weight $w_c$ increases. Importantly, it can be seen that in the case of $w_c=1.0$ in Fig. \ref{fig:td3_improve} b) the reward curriculum performs equal or better for nearly all cases, indicating that it indeed manages to learn the task even with strong constraints. In contrast, the baseline does not learn the problem at all in many cases, such as for finger spin, walker run or cheetah run. Therefore, the majority of the episode reward for the cases where the baseline does not learn the task comes from only optimizing the constraints. Overall, the results confirm that the reward curriculum effectively prevents getting stuck in the local minima of only optimizing the constraints, and continues to enable effective task learning. Consistent with this analysis, comparisons of RC-SAC against SAC exhibit similar trends (see Appendix \ref{fig:sac_improvement}), thereby substantiating the efficacy of our proposed approach


To better understand the impact of each component of our method, we perform the following ablation studies. There are two ways in which the reward curriculum can improve performance: i) the agent learned the task in the first phase, which serves as a helpful prior in the second phase, or, ii) the replay buffer with updated rewards helps to stabilize training and to find high reward regions. In order to test the hypotheses we perform experiments on environments where the curriculum was effective, namely on finger spin, walker run, and cartpole swingup. In the first experiment, we compare either resetting the networks or resetting the replay buffer after switching curriculum phases to our method where we keep both. The results are shown in Fig. \ref{fig:td3_resets}. As can be seen, resetting the weights substantially decreases performance while resetting the buffer has little influence. Therefore, we conclude that the main benefit of our method for those environments is to effectively ``pretrain" the networks which help find high-reward regions. Note that the high standard  deviations in the cartpole swingup environment are due to training instabilities, where the agent temporarily unlearns the task. These instabilities occurred later in training, after the curriculum phase had changed, indicating that they are not specific to our method.

The second ablation study focuses on testing the effectiveness of the automatic curriculum switch. We make use of the same environment subset as before, finger spin, walker run, and cartpole swingup. The proposed automatic switch to change the curriculum phase is compared to statically switching at times at T/2, T/3, and T/8. Fig. \ref{fig:td3_switch} illustrates the results. As can be seen, our automatic switch leads to equal or better results than switching at a fixed time while converging to better solutions earlier, thus being more sample-efficient. Table \ref{tab:autoswitch} shows the average switch steps (in [k]) of our method, showing that in most cases phases are switched relatively early, except when learning is challenging. For instance, in finger turn hard it either switches phases late or not at all in the given timesteps, indicating that it did not successfully learn the problem. We observe that the times where the curriculum does not switch phases correspond to an agent not learning the task. Similar training instabilities in some cases as observed before for cartpole swingup also occur in this experiment.

\begin{table}[]
    \centering
 \begin{tabular}{llll}
\toprule
Environment &               $w_c=0.0$ &        $w_c=0.5$ &        $w_c=1.0$ \\
\midrule
fish-upright      &    118 ± 17 &     101 ± 14 &     102 ± 7 * \\
finger-turn\_hard  &  1224 ± 580 &    735 ± 571 &  1220 ± 730 † \\
cheetah-run       &    270 ± 29 &     305 ± 50 &    293 ± 33 * \\
fish-swim         &    252 ± 60 &     273 ± 78 &    265 ± 75 * \\
walker-walk       &    192 ± 18 &     190 ± 47 &      186 ± 30 \\
walker-stand      &   122 ± 4 † &     128 ± 10 &      122 ± 11 \\
walker-run        &    321 ± 64 &     279 ± 67 &      290 ± 48 \\
pendulum-swingup  &  331 ± 76 * &  313 ± 100 * &    488 ± 54 * \\
reacher-hard      &   314 ± 144 &    300 ± 171 &     316 ± 146 \\
reacher-easy      &    104 ± 44 &     124 ± 73 &       93 ± 39 \\
finger-spin       &   669 ± 670 &     255 ± 71 &      303 ± 95 \\
ball\_in\_cup-catch &   396 ± 368 &     194 ± 80 &    227 ± 99 * \\
cartpole-balance  &      83 ± 3 &       84 ± 1 &        82 ± 2 \\
cartpole-swingup  &    124 ± 15 &      122 ± 8 &       120 ± 8 \\
finger-turn\_easy  &    406 ± 70 &     359 ± 50 &      331 ± 73 \\
\bottomrule
\end{tabular}
    \caption{Number of steps [k] when the curriculum phase switches. In certain cases, the second phase was never initiated which is shown by * if $\frac{3}{4}$ and † is $\frac{2}{4}$ runs switched to the second phase. The few times when the curriculum did not switch phases corresponded to unsuccessful initial training.}
    \label{tab:autoswitch}
\end{table}

\begin{figure*}[t]
  \centering
  \includegraphics[width=\textwidth]{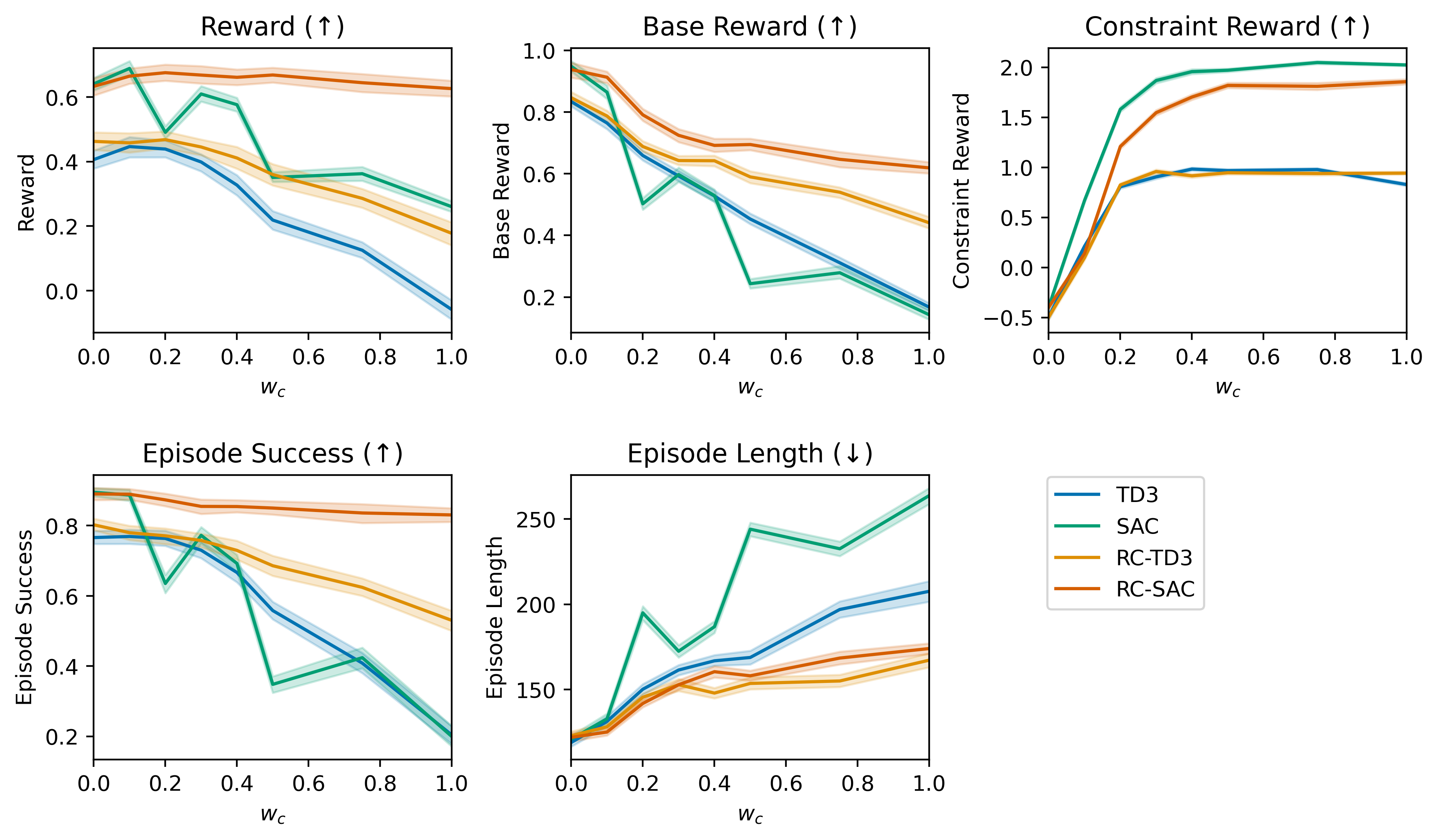}
  \caption{Performance of TD3, SAC, RC-TD3 and RC-SAC for the mobile robot environment. Results are averaged over the last $50,000$ training timesteps and 4 seeds. Both RC-TD3 and RC-SAC consistently outperform TD3 and SAC across all metrics, indicating the effectiveness of a reward curriculum for constraint environments. Interestingly, RC-SAC and SAC manage to better optimize the constraints compared to the TD3 versions. Overall, RC-SAC performs the best and manages to retain high success rates across all constraint weights.}
  \label{fig:mb_wc}
\end{figure*}

\subsection{Mobile Robot}

While there might be cases in which a non-intuitive reward subset to take as base reward $r_b$ is best, we found that simply using the terms that encourage the agent to only learn the task works best (See Appendix \ref{app:base_rewards} for more details). Specifically, we chose the base reward as $r_b = r_g + w_p r_p$, and all other reward terms become part of the constraint reward, thus $r_c=w_c (r_a+r_x+r_v)$. The experiments are repeated for the constraint weights $w_c=\{0.0, 0.1, 0.2, 0.3, 0.4, 0.5, 0.75, 1.0\}$ and train for $T=1,000,000$ steps. We choose these values as we expect more refined changes for smaller $w_c$. In the case of TD3, we start with $\sigma=0.9$ and anneal it to $\sigma=0.1$ over the first T/2 frames for additional exploration.

We present the results of TD3, SAC, RC-TD3, and TC-SAC, averaged over the last $50,000$ training steps, as a function of the selected constraint weights in Fig. \ref{fig:mb_wc}. To ensure comparability, the reward is calculated using a fixed $w_c = 0.1$, while the other terms are independent of $w_c$. Consistent with observations in the DeepMind Control Suite, the reward curriculum exhibits increased effectiveness with growing $w_c$. While the constraint rewards $r_c$ are relatively similar between the reward curriculum versions and their baseline, the curriculum versions more effectively learn the task, indicated by an increased base reward $r_b$. Moreover, SAC and RC-SAC outperform TD3 and RC-TD3 in terms of initial success rate and constraint optimization. Interestingly, RC-SAC achieves a substantially higher success rate, particularly for large $w_c$, surpassing RC-TD3 and yielding the best results in this environment. One reason for this could be that RC-SAC on average switched phases at $477 \pm 96$ while RC-TD3 switched at $170 \pm 44$, indicating that switching later might be beneficial. In addition, the results demonstrated that optimizing the constraints indeed conflicts with task performance, as they exhibit opposing trends. Overall, the value $w_c=0.2$ seems to lead to the highest reward for both RC-TD3 and RC-SAC. These findings corroborate our earlier results and demonstrate the efficacy of our approach in achieving higher rewards compared to the baselines, especially in the presence of strong constraints.

\section{Related Work}
Reward shaping is a well-established technique to facilitate sample-efficient learning, with a history spanning decades \citep{dorigo1994robot, randlov1998learning}. The most prevalent approach is potential-based reward shaping, which preserves the policy ordering \citep{ng1999policy}. Building on this foundation, recent work has explored the concept of automatic reward shaping, where the shaping process is learned and adapted during training. For instance, \citet{hu2020learning} formulate reward shaping as a bi-level optimization problem, where the high-level policy optimizes the true reward and adaptively selects reward shaping weights for the lower-level policy. This approach has been successfully applied to various tasks with sparse rewards, effectively creating an automatic reward-shaping curriculum. Another line of research focuses on building automatic reward curricula through intrinsic motivation, encouraging exploration of unexpected states \citep{bellemare2016unifying, pathak2017curiosity, burda2018exploration, shyam2019model}. However, these methods primarily modify the shaping terms while keeping the ground-truth rewards fixed.

Although less prevalent, several works have employed reward curricula in reinforcement learning (RL) using subsets of the true reward. For instance, \citet{hwangbo2019learning} utilize an exponential reward curriculum, initially focusing on locomotion and gradually increasing the weight of other cost terms to refine the behavior. Similarly, \citet{leyendecker2022deep} observe that RL agents optimized on complex reward functions with multiple constraints are highly sensitive to individual reward weights, leading to local optima when directly optimizing the full reward. They successfully learn a policy by gradually increasing constraint weights based on task success. Furthermore, \citet{pathare2024tactical} employ a three-stage reward curriculum, incrementally increasing complexity and realism by incorporating additional terms. However, they find reward curriculum learning largely ineffective. To the best of our knowledge, we are the first to systematically investigate a two-stage curriculum for complex rewards, leveraging sample-efficient experience transfer between phases to mitigate reward exploitation.

\section{Limitations}
While the proposed reward curriculum shows promising results, there are still remaining open questions. For instance, in some environments it happens that a successful policy in the first phase deteriorates in performance in the second one. Future work could investigate more stable phase switches while retaining enough flexibility to effectively optimize constraints.
Another issue is that currently our method depends on the specific value selected as lower threshold for when an actor sufficiently fits a $Q$-function. This greatly impacts when and if phases are switched, and thus also the final performance.
Furthermore, we do not evaluate our method in experiments using real robots. While the essence of the problem should be comparable, it might bring in new aspects that are challenging to anticipate in simulation.

\section{Conclusion} 
\label{sec:conclusion}
In this work, we present a two-stage reward curriculum, where we first train an RL agent with on an easier subset of rewards and switch to training using the full reward after a certain time. We introduce a flexible replay buffer that adaptively changes rewards to reuse samples from one phase in the next. Further, we provide a mechanism to automatically change phases.
In extensive experiments we show that our method outperforms the baseline only trained on the full reward in almost all cases. It is especially successful for high constraint weights and environments where the constraints substantially hinder learning the task. We believe that this work contributes towards developing more stable RL methods to effectively learn challenging objectives.
For future work, methods to automatically detect a feasible reward subset for the first phase could be developed and further evaluations in real-world experiments are required.

\section*{Acknowledgments}
The computations were enabled by resources provided by the National Academic Infrastructure for Supercomputing in Sweden (NAISS), partially funded by the Swedish Research Council through grant agreement no. 2022-06725. This work is supported by the Vinnova project AIHURO (Intelligent human-robot collaboration) and the ITEA project ArtWork.

\bibliographystyle{plainnat}
\bibliography{references}

\clearpage
\appendix

\subsection{Mobile robot base reward}\label{app:base_rewards}
\begin{figure}[h]
  \centering
  \includegraphics[width=\columnwidth]{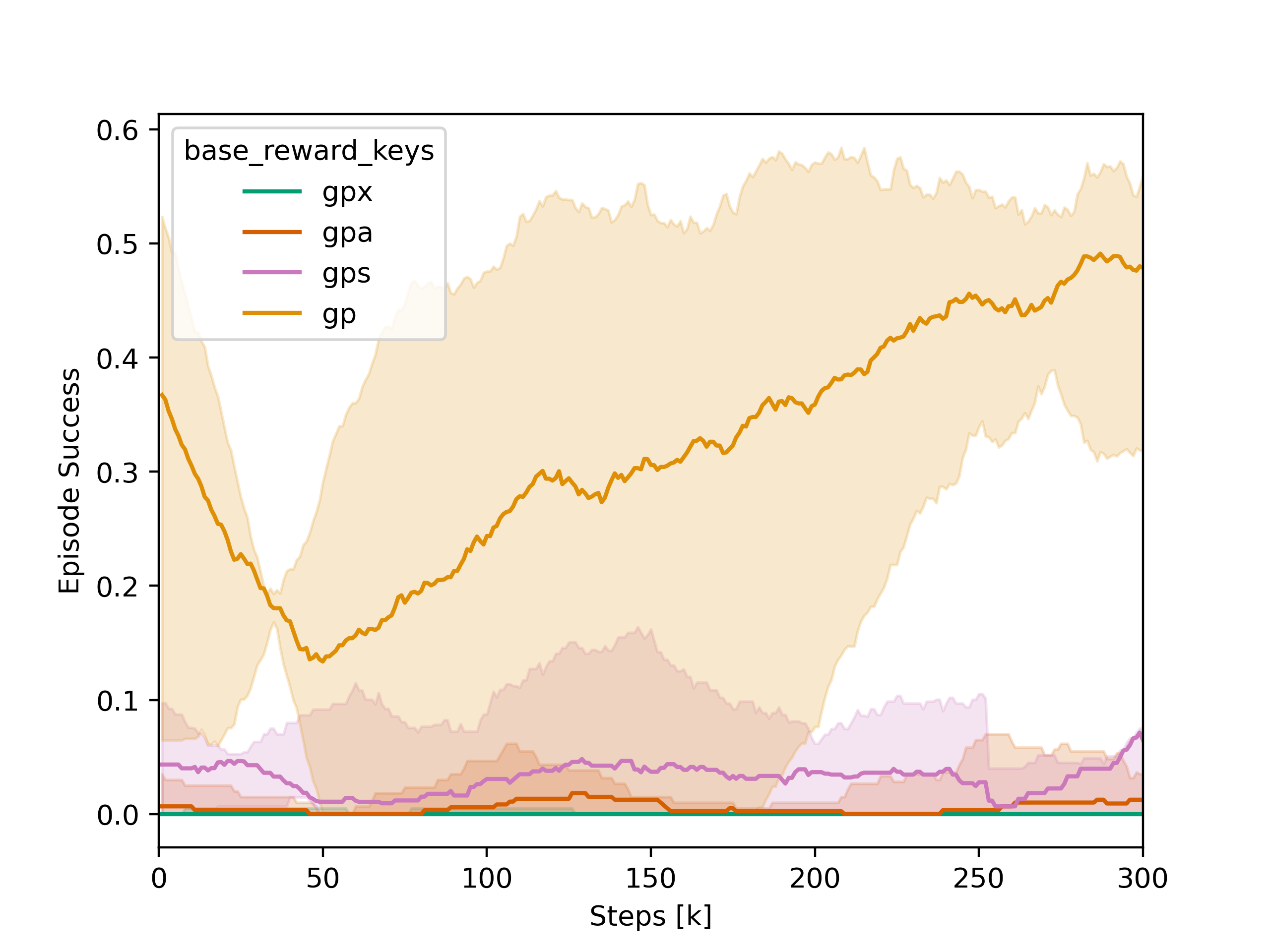}
  \caption{Success rate for the mobile robot during training using TD3 with different initial reward subsets for $w_c=0.5$. As it can be seen, adding any constraint lead to worse performance than only taking the reward terms that intuitively contribute to learning the task.}
  \label{fig:td3_mb_base_rwd}
\end{figure}

An important assumption underlying our approach is that training on a subset of rewards is easier than training on the full reward function. However, while in the case of the DeepMind control suite it was relatively clear which initial reward to use, it might be that some subset of initial rewards $r_b$ converges faster than others. To empirically validate this, we perform experiments on different subsets of the shaped reward function. Specifically, we consider the following subsets: 
\begin{align*}
    r_{b}\in \{r_{\text{full}}, r_{\text{gp}}, r_{\text{gpc}}, r_{\text{gpv}}, r_{\text{gpa}}, r_{\text{gpx}}\}
\end{align*} where
\begin{align*}
\begin{split}
    r_{\text{gp}} &= r_g + r_p, \\
    r_{\text{gpv}} &= r_g + r_p + r_v, \\
    r_{\text{gpa}} &= r_g + r_p + r_a, \\
    r_{\text{gpx}} &= r_g + r_p + r_x.
\end{split}
\end{align*}
We train for $T=300\,000$ environment steps on the randomized maps in Figure \ref{fig:maps} over 4 seeds for each subset with $w_c=0.5$. Fig. \ref{fig:td3_mb_base_rwd} shows the task success rate for each subset over the training steps. It can be concluded that non of the constraints seems to help learning the task and in fact make learning more challenging. Therefore, this confirms that only training on the reward terms that clearly contribute to learning the task is the most beneficial subset for this environment.

\subsection{Additional Results}\label{app:add_results}
\begin{figure*}[h]
  \centering
  \includegraphics[width=0.7\textwidth]{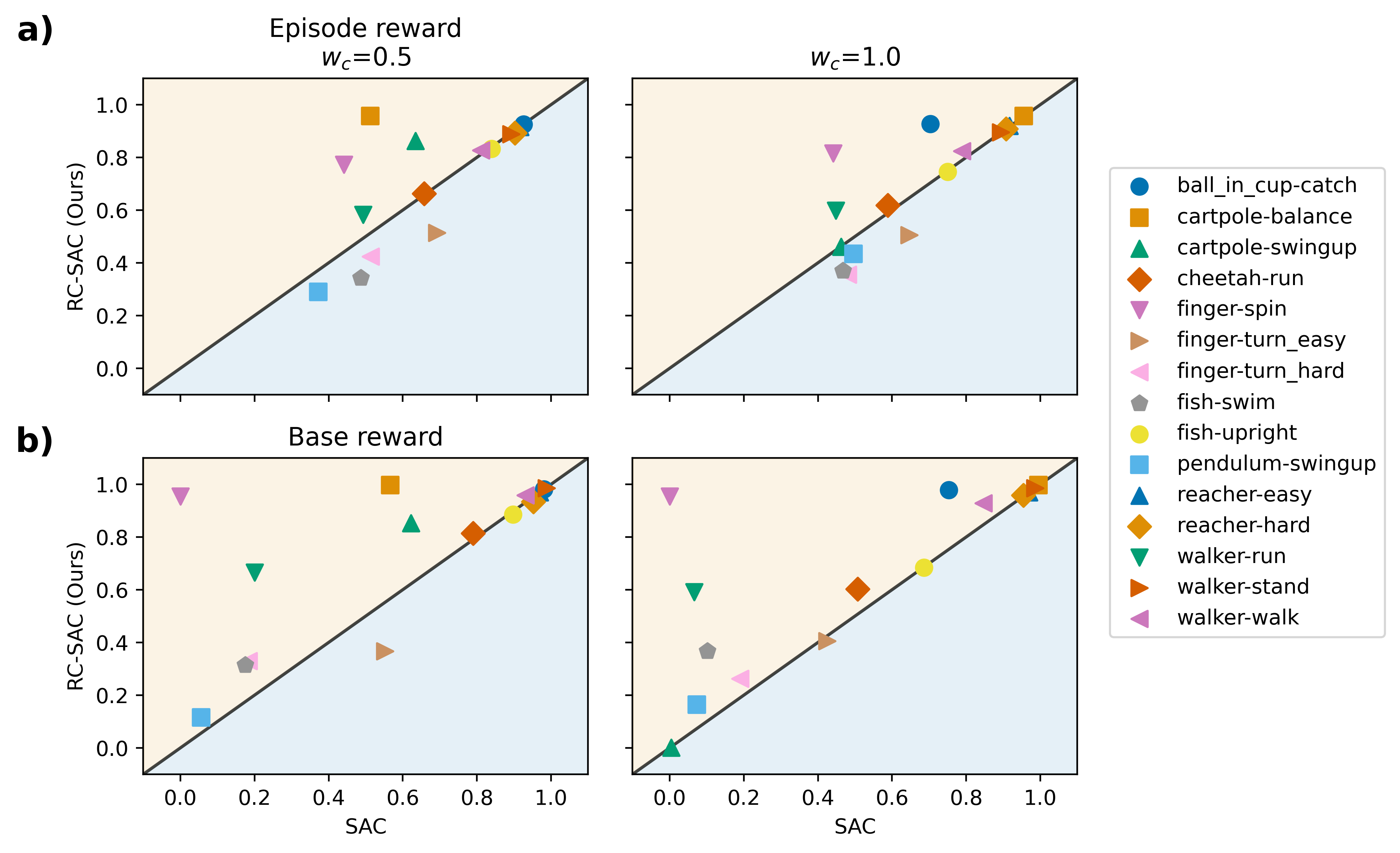}
  \caption{a) Episode reward averaged over 4 random seeds for RC-SAC on the x axis and SAC on the y axis. b) Base reward for the same setup as in before, which corresponds to task performance without considering constraints. Overall, the outcomes are very similar to the ones observed for RC-TD3 and TD3.}
  \label{fig:sac_improvement}
\end{figure*}

Furthermore, we show results obtained with RC-SAC compared to the baseline SAC. A comparison of episode rewards can be seen in Fig. \ref{fig:sac_improvement}. Detailed values can be found in Table \ref{tab:sac_dm}. The trends are similar to RC-TD3, improvements of RC-SAC over SAC become more pronounced for higher $w_c$ and are more relevant for some environments than others. Importantly, the environments where RC-SAC is effective correspond to the ones of RC-TD3. Detailed values of the mean and standard deviations can be seen in Table \ref{tab:td3_dm}. 

\begin{table}[]
    \centering
\begin{tabular}{lllll}
\toprule
\multicolumn{1}{c}{} & \multicolumn{2}{c}{$w_c=0.5$} & \multicolumn{2}{c}{$w_c=1.0$}\\
& SAC & RC-SAC & SAC & RC-SAC \\
 \midrule
walker-walk & 622 ± 63 & \underline{655 ± 38} & 578 ± 99 & \underline{648 ± 96} \\
walker-stand & \underline{785 ± 16} & 777 ± 45 & 787 ± 16 & \underline{792 ± 15} \\
walker-run & -15 ± 51 & \textbf{165 ± 35} & -103 ± 42 & \textbf{196 ± 45} \\
fish-upright & \underline{680 ± 146} & 666 ± 194 & \underline{500 ± 387} & 492 ± 391 \\
pendulum-swingup & \underline{-255 ± 478} & -419 ± 561 & \underline{-8 ± 254} & -131 ± 467 \\
fish-swim & \underline{-26 ± 179} & -314 ± 286 & \underline{-65 ± 119} & -262 ± 296 \\
reacher-hard & \underline{804 ± 142} & 785 ± 192 & 814 ± 135 & \underline{817 ± 104} \\
reacher-easy & \underline{834 ± 101} & 831 ± 97 & 834 ± 151 & \underline{839 ± 105} \\
finger-turn\_hard & \underline{27 ± 369} & -152 ± 610 & \underline{-39 ± 419} & -288 ± 468 \\
ball\_in\_cup-catch & 851 ± 19 & \underline{851 ± 19} & 407 ± 781 & \underline{853 ± 19} \\
finger-turn\_easy & \underline{384 ± 492} & 28 ± 539 & \underline{293 ± 488} & 11 ± 576 \\
finger-spin & -118 ± 2 & \textbf{545 ± 82} & -118 ± 2 & \textbf{630 ± 55} \\
cartpole-balance & 25 ± 891 & \underline{914 ± 2} & 911 ± 10 & \underline{914 ± 2} \\
cartpole-swingup & 268 ± 690 & \underline{725 ± 2} & \underline{-76 ± 4} & -79 ± 2 \\
cheetah-run & 315 ± 64 & \underline{325 ± 81} & 177 ± 79 & \underline{236 ± 42} \\
\bottomrule
\end{tabular}
    \caption{Mean and standard deviation of the episode reward for SAC and RC-SAC for the DeepMind control suite over 4 random seeds. Bold values indicate that one method was clearly better (i.e. the standard deviations did not overlap) and underlines indicate that the mean value was higher. }
    \label{tab:sac_dm}
\end{table}

\begin{table*}[]
    \centering
\begin{tabular}{lllllll}
\toprule
\multicolumn{1}{c}{} & \multicolumn{2}{c}{$w_c=0.0$} & \multicolumn{2}{c}{$w_c=0.5$} & \multicolumn{2}{c}{$w_c=1.0$}\\
 & TD3 & RC-TD3 & TD3 & RC-TD3 & TD3 & RC-TD3 \\
 \midrule
fish-upright & \underline{60 ± 95} & 12 ± 122 & \underline{508 ± 253} & 381 ± 249 & 626 ± 176 & \underline{634 ± 174} \\
finger-turn\_hard & \underline{-788 ± 319} & -828 ± 260 & -810 ± 234 & \underline{-642 ± 355} & -758 ± 298 & \underline{-630 ± 396} \\
finger-turn\_easy & \underline{-700 ± 405} & -713 ± 391 & -697 ± 384 & \underline{-640 ± 362} & -732 ± 335 & \underline{-579 ± 361} \\
cheetah-run & -348 ± 342 & \underline{-155 ± 164} & \underline{-77 ± 1} & -169 ± 276 & -77 ± 1 & \underline{92 ± 227} \\
fish-swim & \underline{-817 ± 170} & -850 ± 123 & \underline{-824 ± 142} & -837 ± 142 & \underline{-801 ± 111} & -830 ± 135 \\
walker-walk & \underline{226 ± 82} & 218 ± 86 & 462 ± 321 & \underline{538 ± 123} & 391 ± 447 & \underline{613 ± 98} \\
walker-stand & \underline{300 ± 83} & 263 ± 102 & \underline{780 ± 41} & 753 ± 66 & 783 ± 128 & \underline{799 ± 105} \\
walker-run & -301 ± 72 & \textbf{-154 ± 67} & -3 ± 24 & \textbf{84 ± 57} & -61 ± 5 & \textbf{202 ± 126} \\
pendulum-swingup & \underline{283 ± 277} & 26 ± 455 & -16 ± 232 & \underline{299 ± 443} & -22 ± 219 & \underline{41 ± 475} \\
reacher-hard & \underline{-656 ± 397} & -809 ± 267 & -857 ± 167 & \underline{-823 ± 179} & -845 ± 155 & \underline{-738 ± 302} \\
reacher-easy & \underline{137 ± 401} & -91 ± 537 & \underline{116 ± 600} & 65 ± 634 & -264 ± 689 & \underline{86 ± 749} \\
finger-spin & \underline{191 ± 34} & 14 ± 198 & -80 ± 1 & \textbf{501 ± 75} & -80 ± 1 & \textbf{561 ± 74} \\
ball\_in\_cup-catch & 110 ± 612 & \underline{400 ± 228} & 766 ± 306 & \underline{884 ± 22} & \underline{867 ± 34} & 439 ± 760 \\
cartpole-balance & 704 ± 186 & \underline{779 ± 152} & \underline{758 ± 292} & 756 ± 306 & 798 ± 226 & \underline{847 ± 73} \\
cartpole-swingup & \underline{571 ± 144} & 569 ± 94 & \underline{646 ± 84} & 209 ± 636 & 388 ± 332 & \underline{429 ± 357} \\
\bottomrule
\end{tabular}
    \caption{Mean and standard deviation of the episode reward for TD3 and RC-TD3 for the DeepMind control suite over 4 random seeds. Bold values indicate that one method was clearly better (i.e. the standard deviations did not overlap) and underlines indicate that the mean value was higher. }
    \label{tab:td3_dm}
\end{table*}

\end{document}